# Predicting the Severity of Breast Masses with Data Mining Methods

Sahar A. Mokhtar [1]  and  Alaa. M. Elsayad [2]

[1, 2] Computer and Systems Department, Electronic Research Institute
Dokki, 12622/ Cairo, Egypt
*[sahar, sayad]@eri.sci.eg*

**Abstract**

Mammography is the most effective and available tool for breast cancer screening. However, the low positive predictive value of breast biopsy resulting from mammogram interpretation leads to approximately 70% unnecessary biopsies with benign outcomes. Data mining algorithms could be used to help physicians in their decisions to perform a breast biopsy on a suspicious lesion seen in a mammogram image or to perform a short term follow-up examination instead. In this research paper data mining classification algorithms; Decision Tree (DT), Artificial Neural Network (ANN), and Support Vector Machine (SVM) are analyzed on mammographic masses dataset. The purpose of this study is to increase the ability of physicians to determine the severity (benign or malignant) of a mammographic mass lesion from BI-RADS attributes and the patient's age. The whole dataset is divided for training the models and test them by the ratio of 70:30% respectively and the performances of classification algorithms are compared through three statistical measures; sensitivity, specificity, and classification accuracy. Accuracy of DT, ANN and SVM are 78.12%, 80.56% and 81.25% of test samples respectively. Our analysis shows that out of these three classification models SVM predicts severity of breast cancer with least error rate and highest accuracy.

## 1. Introduction

Breast cancer is a very common and serious cancer for women. It is the second largest cause of cancer deaths among women. Mammography is one of the most used methods to detect this kind of cancer [3, 4]. The value of mammography is that it can identify breast abnormalities with 85-90% accuracy. In literature, radiologists show considerable variation in interpreting a mammography. In such cases, Fine Needle Aspiration Cytology (FNAC) is adopted. But, the average correct identification rate of FNAC is only 90% [5]. It is necessary to develop better identification method to recognize the breast cancer. Computer aided diagnosis can help to reduce the number of false positives and therefore reduce the number of unnecessary biopsies. Machine learning algorithms and artificial intelligence methods have been successfully used to predict the breast cancer by several researchers [6,7]. The objective of these identification techniques is to assign a patient to either a benign group that does not have breast cancer or a malignant group who has strong evidence of having breast cancer.

A study was conducted to demonstrate that machine learning algorithms can help in making correct diagnosis in [8]. Their results show that even the most experienced physician can diagnose properly (79.97%) when compared to the diagnosis made with the help of machine learning and expert system (91.1%). Accordingly, data mining algorithms have been heavily used in the medical field, to include patient diagnosis records to help identify best practices. The difficulties posed by prediction problems have resulted in a variety of problem-solving techniques. For example, data mining methods comprise artificial neural networks and decision trees, and statistical techniques include linear regression and stepwise polynomial regression [9]. However, it is required to compare the accuracy of these algorithms and determine the best one as their presentation is data dependent.

In this study: three of the effective data mining classification techniques; Decision Tree (DT), Artificial neural networks (ANNs), and Support Vector Machine (SVM) are analyzed and evaluated for prediction of the severity of breast masses. These mining algorithms have remarkable ability to derive meaning from complicated or imprecise data. Also, they are able to extract patterns and to detect trends that are too complex to be noticed by either humans or other conventional techniques. The mammographic dataset investigated in this study has been collected at the Institute of Radiology of the University Erlangen-Nuremberg between 2003 and 2006 [10]. However, a few studies applied and compared the performance of data mining and statistical approaches for the diagnosis of mammographic masses data set. A. Keles et al in [11] provided an expert system to diagnose the mammogram masses using neuro-fuzzy rules. In [12], A. Elsayad presented an ensemble of Bayesian networks to classify the same dataset and compared its performance with the one of multilayered neural network classifier. An essential contribution of this study is to benchmark the performance of the three models DT with Chi-squared automatic interaction detection, ANN with pruning parameters and SVM with polynomial kernel for predicting the severity of mammographic masses using different statistical measures including classification accuracy, sensitivity and specificity error rate, true positive rate and false positive rate.

## 2. Material and Methods

2.1 Mammographic Masses Data Set

A radiologist is a physician who analyzes the radiograph to decide if there is a tumor or just normal tissue and whether the existing tumor is malignant (cancerous) or benign (gentle). Due to the variations in mammography interpretations, the problem is gotten ahead to the pathologist. A pathologist is a physician who analyzes cells and tissues under a microscope to determine whether they are malignant or benign. The pathologist's report helps characterize specimens taken during biopsy or other surgical procedures and helps determine treatment. To determine a tumor's histologic grade, a sample of breast cells must be taken from a breast biopsy, lumpectomy or mastectomy. The purpose of this study is to increase the ability of physicians to determine the severity (benign or malignant) of a mammographic mass lesion from BI-RADS attributes and the patient's age. The objective is to reduce the high number of unnecessary breast biopsies. The six BI-RADS reporting categories represent gradations of the likelihood that a cancer exists, from lowest to highest probability. The mammographic mass dataset used here has been collected at the Institute of Radiology of the University Erlangen-Nuremberg between 2003 and 2006 [10]. BI-RADS stands for the Breast Imaging and Reporting Data System and was developed by the American College of Radiology (ACR), in collaboration with multiple other organizations in 1991 to present answers concern about ambiguous mammography reports with indecisive conclusions from radiologists [13].

Table 1: Attributes of mammographic mass dataset

| Attribute | Type | Value | Label | No. of missing values |
|---|---|---|---|---|
| BI-RADS assessment (non-predictive) | Ordinal | 0 | Assessment incomplete | 2 |
| | | 1 | Negative | |
| | | 2 | Benign findings | |
| | | 3 | Probably benign | |
| | | 4 | Suspicious abnormality | |
| | | 5 | Highly suggestive of malignancy | |
| Ages | Integer | | Patient's age in years | 5 |
| Mass shape | Nominal | 1 | Round | 31 |
| | | 2 | Oval | |
| | | 3 | Lobular | |
| | | 4 | Irregular | |
| Mass margin | Nominal | 1 | Circumscribed | 48 |
| | | 2 | Microlobulated | |
| | | 3 | Obscured | |
| | | 4 | Ill-defined | |
| | | 5 | Speculated | |
| Mass density | Ordinal | 1 | High | 76 |
| | | 2 | Iso | |
| | | 3 | Low | |
| | | 4 | Fat-containing | |
| Severity (target class) | Binominal | 0 | Benign | |
| | | 1 | Malignant | |

The data set is available by http access of the University of California at Irvine (UCI) machine learning repository [1]. Table 1 shows the mammographic mass dataset which contains the BI-RADS assessment, the patient's age and three BI-RADS attributes together with the ground truth (the severity attribute). The dataset contains 961 sample with class distribution: benign: 516; malignant: 445. There are 162 missing values of different attributes. The values of ordinal attribute represent categories with some intrinsic ranking while they nominal attribute represent categories with no intrinsic ranking in nominal type.

2.2 Decision Tree (DT)

DT models are powerful classification algorithms. They become increasingly more popular with the growth of data mining applications [14, 15]. As the name implies, this model recursively splits data samples into branches to construct a tree structure for the purpose of improving the prediction accuracy. Each tree node is either a leaf node or decision node. All decision nodes have splits, testing the values of some functions of data attributes. Each branch from the decision node corresponds to a different outcome of the test. Each leaf node has a class label attached to it. Fig. 1 shows an illustrating example for binary tree where each decision node has two splits only. CHAID-DT model allows multiple splits on a predictive attribute. This model relies on the Chi-square $\chi^2$ test to determine the best split at each step. CHAID-DT only accepts nominal or ordinal categorical predictive attributes [16]. When predictors are continuous, they are transformed into ordinal ones. Ordinal attribute is ordered set with intrinsic ranking.

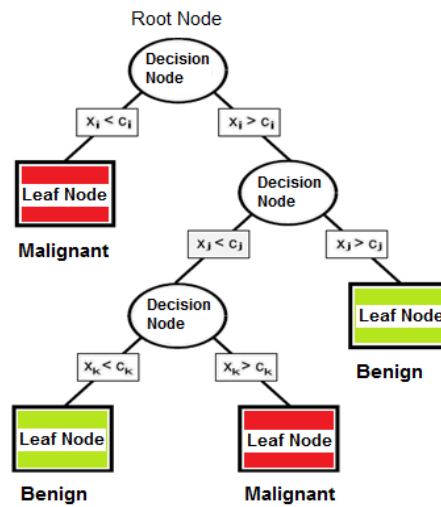

Fig. 1 Example of a binary decision tree.

The CHAID-DT modeling algorithm is as follows [15]:
1. Binning the continuous attribute (if exists) to create a set of categories, where each category is a subrange along the entire range of the attribute. This binning operation permits CHAID-DT model to accept both categorical and continuous inputs, although it internally works only with categorical ones.
2. Analyzing the categories of each attribute to determine which ones can be merged safely to reduce the number of categories.
3. Computing the adjusted p-value for the merged categories by applying Bonferroni adjustments.
4. Searching for the split point with the smallest adjusted p-value (probability value, which can be related to significance) to find the best split.

In step 2, the algorithm merges values that are judged to be statistically homogeneous (similar) with respect to the target attribute and maintains all other values that are heterogeneous (dissimilar). If the p-value is greater than specified parameter αmerg then the algorithm merges the pair of categories into a single one. The value of αmerg must be greater than 0 and less than or equal to 1. To prevent any merging of categories, specify a value of 1. In step 4, each predictive attribute is evaluated for its association with the target attribute, based on the adjusted p-value of the statistical test of association. The predictive attribute with the strongest association, indicated by the smallest adjusted p-value, is compared to a pre-specified split threshold αsplit. If the adjusted p-value is less than or equal to αsplit, that attribute is selected as the split attribute for the current node. After the split is applied to the current node, the child nodes are examined to see if they warrant splitting by applying the merge/split process to each in turn. Processing proceeds recursively until one or more stopping rules are triggered for every unsplit node, and no further splits can be made.

In this study, the target attribute is of categorical type (malignant or benign). The Likelihood ratio has been used to compute the chi-square statistic. The algorithm forms a contingency (count) table using the classes of the target attribute $y$ as columns and the categories of the predictive attribute $x$ as rows. The expected cell frequencies under the null hypothesis of independence are estimated. The observed cell frequencies and the expected cell ones are used to calculate the chi-squared statistic and the *p-value*.

$$G^2 = \sum_{j=1}^{J}\sum_{i=1}^{I} n_{ij} \ln\left(\frac{n_{ij}}{\widehat{m}_{ij}}\right) \quad (1)$$

where $n_{ij} = \sum_n f_n I(x_n = i \wedge y_n = j)$ is the observed cell frequency and $\widehat{m}_{ij}$ is the expected cell frequency for cell $(x_n = i, y_n = j)$, and the *p-value* is computed as follows:

$$p = \Pr(\chi_d^2 > G^2) \quad (2)$$

The CHAID-DT model is fast, builds "wider" decision trees as it is not constrained to make binary splits; making it very popular in data mining research.

## 2.3 Artificial Neural Network (ANN)

ANN is a powerful function approximator for prediction and classification problems [17,18]. Multilayer perceptron is debatably the most commonly used and well-studied ANN architecture. It is organized into layers of neurons as input, output and hidden layers as shown in Fig. 2. There is at least one hidden layer, where the actual computations of the network are processed. Each neuron in the hidden layer sums its input attributes $x_i$ after multiplying them by the strengths of the respective connection weights $w_{ij}$ and computes its output $y_j$ using activation function (AF) of this sum. AF may range from a simple threshold function, or a sigmoidal, hyperbolic tangent, or radial basis function.

$$y_i = f(\sum w_{ij} x_i), \quad (3)$$

Where $f$ is the activation function.

The available dataset is normally divided into training and test subsets. Back-propagation (BP) is a common training technique. It works by presenting each input sample to the network where the estimated output is computed by performing weighted sums and transfer functions. The sum of squared differences between the desired and actual values of the output neurons $E$ is defined as

$$E = \frac{1}{2}\sum_j (y_{dj} - y_j)^2 \quad (4)$$

where $y_{dj}$ is the desired value of output neuron $j$ and $y_j$ is the actual output of that neuron.

Weights $w_{ij}$ in Equation (3), are adjusted to reduce the error $E$ of Equation (4) as fast, quickly as possible.

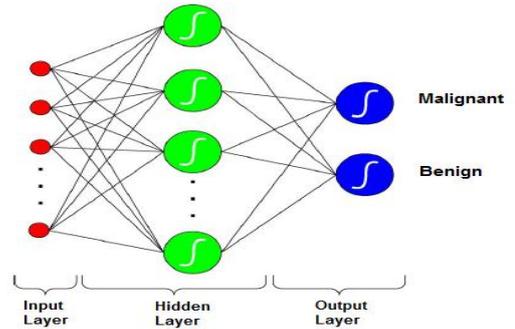

Fig. 2 The structure of multilayer perceptron neural network.

BP applies a weight correction to reduce the difference between the network outputs and the desired ones; i.e., the neural network can learn, and can thus reduce the future errors. The performance of MLPNN depends on network parameters, the network weights and the type of transfer functions used [19]. When using an MLPNN, three important issues need to be addressed; the selection of data samples for network training, the selection of an appropriate and efficient training algorithm and determination of network size. New algorithms for data portioning [20] and efficient training with faster convergence properties and less computational requirements are being developed [21]. However, the third issue is a more difficult problem to solve. It is necessary to find a network structure small enough to meet certain performance specifications. Pruning methods for improving the input-side redundant connections were also developed that resulted in smaller networks without degrading or compromising their performance [22].

2.4 Support Vector Machine (SVM)

SVM is a supervised machine learning technique, which is based on the statistical learning theory. It was firstly proposed by Cortes and Vapnik from his original work on structural risk minimization in [23] and modified by Vapnik in [24]. The algorithm of SVM is able to create a complex decision boundary between two classes with good classification ability. The basic idea is to map the data into a higher dimensional feature space via a nonlinear mapping kernel function chosen a priori (Fig. 3), and constructs a hyperplane, which splits class members from non-members (Fig. 4). SVMs introduce the concept of 'margin' on either side of a hyperplane that separates the two classes. Maximizing the margins and thus creating the largest possible distance between the separating hyperplane and the samples on either side, is proven to reduce an upper bound on the expected generalization error. SVM may be considered as a linear classifier in the feature space. On the other side it becomes a nonlinear classifier as a result of the nonlinear mapping from the input space to the feature one [25, 26]. For linearly separable classes, SVM divides these classes by finding the optimal (with maximum margin) separating hyperplane. Optimal hyperplane can be found by solving a convex quadratic programming (QP) problem [15].

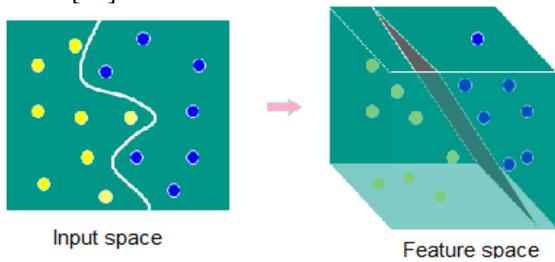

Fig. 3 Mapping kernel.

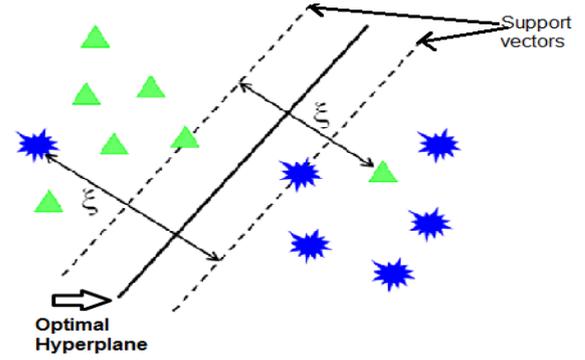

Fig. 4 Optimal hyperplane.

Once the optimum separating hyperplane is found, data samples that lie on its margins are known as support vectors. The solution to this optimization problem is a global one. For linearly decision space, suppose the training subset consists of $n$ samples $(x_1,y_1),\ldots,(x_n,y_n)$, $x \in R^p$ and $y \in \{+1,-1\}$ i.e. the data contains only two classes. The separating hyperplane can be written as

$$D(x_i) = wx_i + b \qquad (5)$$

where the vector $w$ and constant $b$ are learned from a training subset of linearly separable samples.

The solution of SVM is equivalent to solve a linearly constrained quadratic programming problem as Equation (6) for both targets $y = 1$ and $-1$.

$$y_i = wx_i + b \geq 1, \ i=1,\ldots,n. \qquad (6)$$

As mentioned before, samples that provide the above formula in case of equality are referred as support vectors. SVM classifies any new sample using these support vectors. On the other hand, margins of the hyperplane follow the subsequent inequality

$$\frac{y_i \times D(x_i)}{\|w\|} \geq \Gamma, \ i=1,\ldots,n. \qquad (7)$$

The norm of the $w$ has to be minimized in order to maximize the margin $\Gamma$. In order to lessen the number of solutions to the norm of $w$, the following equation is assumed

$$\Gamma \times \|w\| = 1 \qquad (8)$$

Then the algorithm tries to minimize the value of $1/2\|w\|^2$ subject to the condition (6). In the case of non-separable samples, slack parameters $\xi$ are added into the condition as follows.

$$y_i(wx_i + b) \geq 1 - \xi, \ \xi \geq 0, \quad \forall i \qquad (9)$$

And the value that we want to minimize becomes:

$$C\sum_{i=1}^{n} \xi_i + 1/2\|w\|^2. \qquad (10)$$

$C$ is a regularization parameter to determine the level of tolerance the model has, with larger $C$ values allowing larger deviations from the optimal solution. This parameter is optimized to balance the classification error with the complexity of the model. There is a family of kernel functions that may be used to map input space into feature space. They range from simple linear and polynomial mappings to sigmoids and radial basis functions. Once a hyperplane has been created, the kernel function is used to map new samples into the feature space for classification. This mapping technique makes SVM dimensionally independent, whereas other machine learning techniques are not. In this study, we used the polynomial function kernel to map the input space into the higher dimensional feature space. Polynomial kernels can be controlled by adjusting the complexity of the mapping $d$, the coefficient $r$ and weight value $\gamma$ in the kernel function (Equation 11) [27].

$$K(x_i, x_j) = \left(\gamma x_i^T \cdot x_j + r\right)^d, \quad \gamma > 0. \quad (11)$$

where $\gamma$, $r$ and $d$ are the kernel parameters

Consequently, there are four parameters need to be optimized in the polynomial kernel-SVM model; $C$, $\gamma$, $r$ and $d$.

### 2.5 Statistical performance measure

The performance of each classification model is evaluated using three statistical measures; classification accuracy, sensitivity and specificity. These measures are defined using True Positive (TP), True Negative (TN), False Positive (FP) and False Negative (FN). A true positive decision occurs when the positive prediction of the classifier coincided with a positive prediction of the physician. A true negative decision occurs when both the classifier and the physician suggest the absence of a positive prediction. False positive occurs when the system labels a benign case; a negative one as a positive one (malignant). Finally, false negative occurs when the system labels a positive case as negative (benign). Classification accuracy is defined as the ratio of the number of correctly classified cases and is equal to the sum of TP and TN divided by the total number of cases N:

$$\text{Accuracy} = \frac{TP + TN}{N} \quad (12)$$

Sensitivity refers to the rate of correctly classified positive and is equal to TP divided by the sum of TP and FN. Sensitivity may be referred as a True Positive Rate

$$\text{Sensitivity} = \frac{TP}{TP + FN} \quad (13)$$

Specificity refers to the rate of correctly classified negative and is equal to the ratio of TN to the sum of TN and FP. False Positive Rate equals (100-specificity):

$$\text{Specificity} = \frac{TN}{TN + FP} \quad (14)$$

## 3. Results and discussion

The whole dataset is divided for training the models and test them by the ratio of 70:30% respectively. The training set is used to estimate each model parameters, while the test set is used to independently assess the individual models. Fig. 5 shows components of the proposed stream of predicting the severity of breast cancer. The stream is implemented in SPSS Clementine data mining workbench using Intel core 2 Dup CPU with 2.1 GHz. Clementine uses client/server architecture to distribute requests for resource-intensive operations to powerful server software, resulting in faster performance on larger datasets [2]. The software offers many modeling techniques, such as prediction, classification, segmentation, and association detection algorithms. The components of the data mining streams are as follows:

*Mammographic mass dataset node* is connected directly to SPSS file that contains the source data. The dataset was explored for incorrect, inconsistent. Only, the age attribute is normalized and no preprocessing for other attributes. They are ordinal and nominal data types.

*Type node* specifies the field metadata and properties that are important for modeling and other work in Clementine. These properties include specifying a usage type, setting options for handling missing values, as well as setting the role of an attribute for modeling purposes.

*Data audit node* provides a comprehensive first look at the attribute values in an easy-to-read matrix that can be sorted and used to generate full-size graphs. This node computes different descriptive statistics and histogram of each attribute in the dataset. The audit report lists the percentage of complete records for each field, along with the number of valid, null, and blank values. The missing values may be imputed using one of allowable methods for specific fields as appropriate, and then generate a SuperNode to apply these transformations. SuperNodes is used to group multiple nodes which are required for missing value imputation into a single node by encapsulating sections of a data stream. This research substitutes the values predicted by a model based on the C&RT algorithm [2]. For each field imputed using this method, there will be a separate C&RT model, along with a Filler node that replaces blanks and nulls with the value predicted by the model. A Filter node is then used to remove the prediction fields generated by the model.

*Partition node* is used to generate a partition field that splits the data into separate subsets for the training and test the models. In this study, the dataset was partitioned by the ratio 70:30% for training and test subsets respectively.

*CHAID classifier node* is to train the decision tree using the significance of a statistical test as a criterion. The chi-square statistic is computed using the likelihood ratio to merge values that are judged to be statistically homogeneous with respect to the target attribute. After merging similar categories, the algorithm selects the best predictive attribute to form the first branch in the decision tree, such that each child node is made of a group of homogeneous values of the selected attribute. The best values of α merge and α split are found to be 0.1 for both and the maximum allowable depth is set to 5. The resulting CHAID accuracies for training and testing are 81.43% and 78.12% respectively. This model is very fast; it takes below one second to build the model and easy to interpret (Fig. 6).

*ANN classifier node* is trained using the pruning method. It begins with a large network and removes the weakest neurons in the hidden and input layers as training proceeds. The network is trained on the training subset and the stopping criterion is set based on time (one minute). Using the mammographic masses dataset, the resulting structure consists of four layers; one input, two hidden layers and one at the output with 12, 30, 18 and 1 neuron respectively. The prediction accuracies of training and test samples are 81.13% and 80.56% respectively. These results assured the conclusion of [28] where they empirically stated that given the right size and the structure, the ANN is capable of learning arbitrarily complex nonlinear functions to arbitrary accuracy levels.

*SVM classifier node* is used to train SVM with polynomial kernel. There are four parameters that need to be optimized; $C, \gamma, r$ and $d$. The value of the regularization parameter $C$ should be set between 1 and 10 inclusive; increasing the value improves the prediction accuracy for the training data, but this can also lead to overfitting. Using trial and error we found that the best values for these parameters are 10, 1, 0.1 and 4 for $C, \gamma, r$ and $d$ respectively. These values result in 83.66% and 81.25% prediction accuracies for training and test subsets respectively. It takes only 35 seconds to build the model.

*Filter, Analysis and Evaluation nodes* are used to select and rename the classifier outputs in order to compute the performance statistical measures and to graph the evaluation charts.

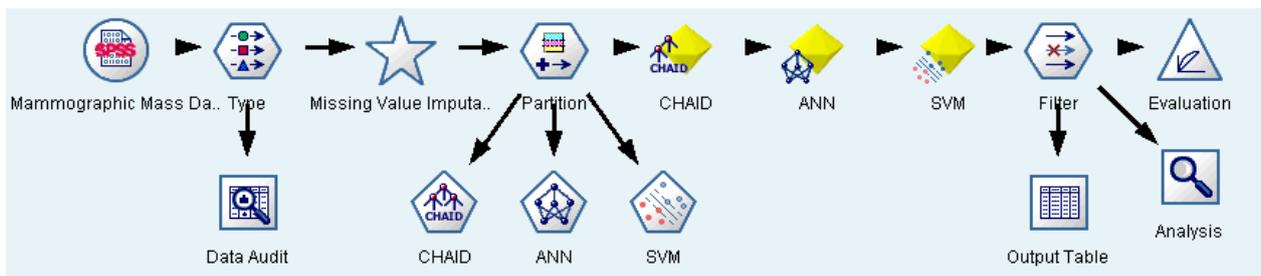

Fig. 5 Stream of classification models to predict the severity of breast masses.

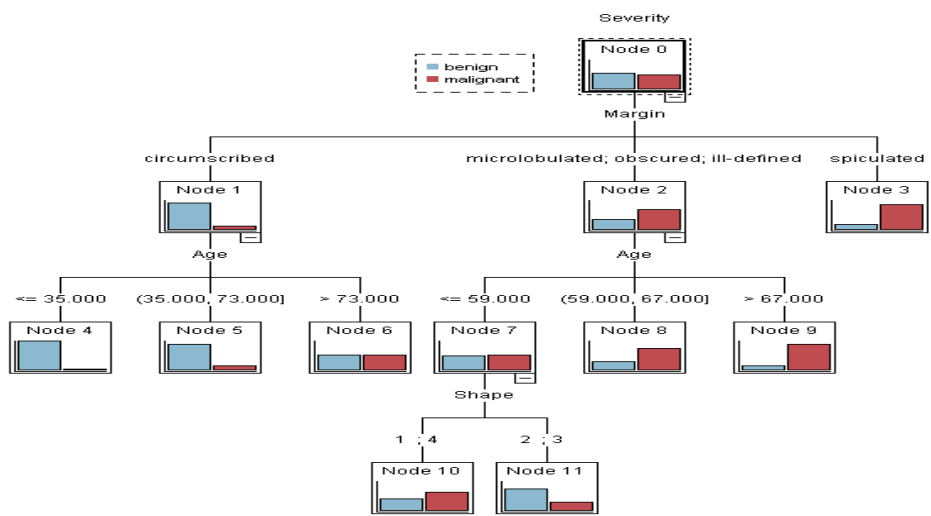

Fig. 6 The graphical representation of CHAID-DT model; this tree is sufficient for correct prediction of 81.43% of the training samples and 78.12% of the test ones.

Table 2 shows the computed confusion matrix, each cell contains the raw number of samples classified for the corresponding combination of desired and actual model outputs. Table 3 presents the values of the statistical parameters (sensitivity, specificity and total classification accuracy) of the predictive models. Sensitivity and Specificity approximates the probability of the positive and negative labels being true. These results show that the sensitivity, specificity and classification accuracy of SVM model are better than those of the other classifiers.

Table 2: Confusion matrices of different models of training and test data partitions

| Model | Desired Output | Training data | | Test data | |
|---|---|---|---|---|---|
| | | Benign | Malignant | Benign | Malignant |
| CHAID | Benign | 283 | 81 | 108 | 44 |
| | Malignant | 44 | 265 | 19 | 117 |
| ANN | Benign | 284 | 80 | 166 | 36 |
| | Malignant | 47 | 262 | 20 | 116 |
| SVM | Benign | 298 | 66 | 119 | 33 |
| | Benign | 44 | 265 | 21 | 115 |

Table 3: The values of the statistical measures for different models of training and test data partitions

| Model | Partition | Measures | | |
|---|---|---|---|---|
| | | Accuracy | Sensitivity | Specificity |
| CHAID | Training | 81.43% | 85.76% | 77.75% |
| | Test | 78.13% | 86.03% | 71.05% |
| ANN | Training | 81.13% | 84.79% | 78.02% |
| | Test | 83.43% | 85.29% | 82.18% |
| SVM | Training | 83.66% | 85.76% | 81.87% |
| | Test | 81.25% | 84.56% | 78.29% |

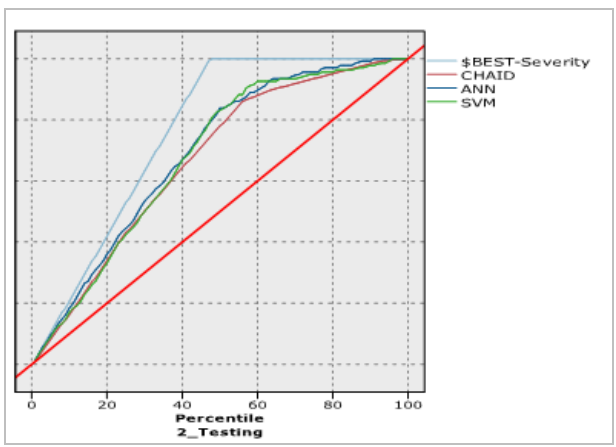

(a)

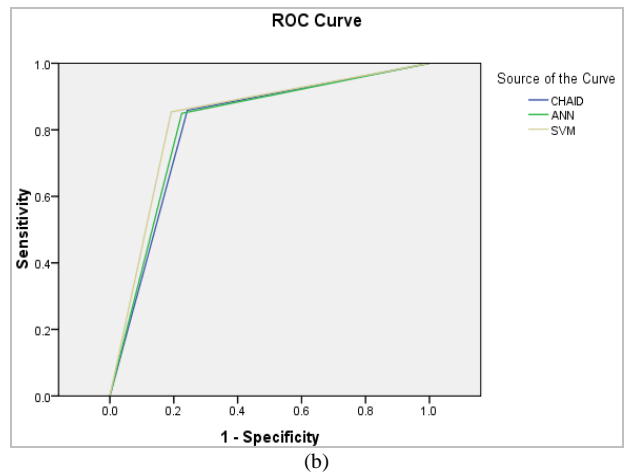

(b)

Fig. 7 ROC curve and gain chart for class severity = 1 of all classifiers. (a) gain chart; (b) ROC curve

Gain and Receiver Operating Characteristic (ROC) curves have been used to compare the performances of different predictive models. The gain curves summarize the utility that can be expected by using the respective predictive models, as compared to using baseline information only. Fig. 7a shows the cumulative gain curves of the three models for test samples. The higher lines indicate better models, especially on the left side of the chart. ROC procedure is a common way to evaluate the performance of classification models in which the class attribute has two categories by which samples are classified. It is a plot of the sensitivity against one minus the specificity for different values of the threshold. Fig. 7b shows the ROC curve of the experimental results. Comparison is usually in terms of the area under the curve, which gives a summary of performance over the whole range of values and is independent of the prevalence of the condition unlike the accuracy, which weights the sensitivity and specificity in proportion to their prevalence. ROC measures the probability that for any pair of patients, one of whom with an event and one without, the patient for whom the event has occurred will have a higher predicted probability of the event than the other. Table 4 shows the area under the ROC curve for each predicting model. SVM has the best value among other models with 0.831 of ROC area curve.

Table 4: Area under the ROC curve

| Model | Area |
|---|---|
| CHAID | 0.808 |
| ANN | 0.812 |
| SVM | 0.831 |

## 4. Conclusions

Recent advances in the field of data mining have driven to the emergence of expert systems for medical applications. Many computational tools and algorithms have been recently developed to increase the experiences and the abilities of physicians for taking decisions about different diseases. Normally physician acquires knowledge and experience after analyzing sufficient number of cases. This experience is reached only in the middle of a physician's career. However, for the case of rare or new diseases, experienced physicians are also in the same situation as new comers. Data mining algorithms can help in making correct diagnosis. The objective is not to replace medical professionals and researchers, but to increase their ability when taking decisions about their patients.

In this paper, three different classification models have been analyzed for the prediction of the severity of breast masses. These models are derived from different family namely; decision tree, artificial neural network and support vector machine. These models are optimized using different methods. The decision tree model is constructed using the Chi-squared automatic interaction detection method and pruning method was used to find the optimal structure of artificial neural network model and finally, support vector machine has been built using polynomial kernel.

The mammographic dataset contains BI-RADS assessment, age, three BI-RADS attributes and type of severity. The proposed stream imputes the missing values then trains and optimizes the three models. The performances of the three models have been evaluated using statistical measures, gain and ROC charts. Support vector machine model outperformed the other two models on the prediction of the severity of breast masses. These results recommend that SVM model can be used effectively in real world systems to predict the severity of breast masses.